%% file: main.tex
\documentclass[runningheads]{llncs}
\usepackage{graphicx}
\usepackage{booktabs}

\usepackage{capt-of}
\usepackage{tabu}
\usepackage{stmaryrd}

\usepackage{algorithm}%
\usepackage{algpseudocode}%

\algrenewcommand\alglinenumber[1]{{\sf\footnotesize#1}}
\algrenewcommand\algorithmicrequire{\textbf{Input:}}
\algrenewcommand\algorithmicensure{\textbf{Output:}}

\usepackage{enumerate}
\usepackage{cite}
\usepackage{times}
\usepackage{url}
\usepackage{hypernat}
\usepackage{graphicx}
\usepackage{setspace}
\usepackage{mdwlist}
\usepackage{amsmath}
\usepackage{amssymb}
\usepackage{tabularx}
\usepackage{xspace}
\usepackage{color}

\newcommand{\abr}{\texttt{abr}\xspace}
\newcommand{\br}{\texttt{br}\xspace}

\newcommand{\dt}{\texttt{dt}\xspace}
\newcommand{\etr}{\texttt{etr}\xspace}
\newcommand{\gbr}{\texttt{gbr}\xspace}
\newcommand{\knn}{\texttt{knn}\xspace}
\newcommand{\lm}{\texttt{lm}\xspace}
\newcommand{\rfr}{\texttt{rfr}\xspace}
\newcommand{\rg}{\texttt{rg}\xspace}

\newcommand{\svm}{\texttt{svm}\xspace}
\newcommand{\xgb}{\texttt{xgb}\xspace}

\newcommand{\dnn}{\texttt{dnn}\xspace}

\newcommand{\cD}{\mathcal{D}} %

\usepackage{cleveref}
\crefname{listing}{source code}{source codes}

\newcommand{\ext}{exhaust temperature\xspace}
\newcommand{\expr}{exhaust pressure\xspace}
\newcommand{\no}{NO\xspace}
\newcommand{\co}{CO\xspace}
\newcommand{\et}{engine torque\xspace}

\newcommand{\trainone}{\texttt{train-data-1}\xspace}
\newcommand{\testone}{\texttt{test-data-1a}\xspace}
\newcommand{\testtwo}{\texttt{test-data-1b}\xspace}
\newcommand{\testthree}{\texttt{test-data-2}\xspace}

\usepackage{caption}
\usepackage{subcaption}
\usepackage{wrapfig}
\usepackage[title]{appendix}

\usepackage[normalem]{ulem} %

\title{MaLTESE: Large-Scale Simulation-Driven Machine Learning for Transient Driving Cycles}
\author{Shashi M. Aithal \and  Prasanna Balaprakash }

\institute{
Argonne National Laboratory, Lemont, IL, USA\\
\email{\{aithal,pbalapra\}@anl.gov}\\ 
}

\titlerunning{MaLTESE: Large-Scale Simulation-Driven ML for Transient Driving Cycles}

\begin{document}

\maketitle
\begin{abstract}
Optimal engine operation during a transient driving cycle is the key to achieving greater fuel economy, engine efficiency, and reduced emissions.  In order to achieve continuously optimal engine operation, engine calibration methods use a combination of static correlations obtained from dynamometer tests for steady-state operating points and road and/or track performance data. As the parameter space of control variables, design variable constraints, and objective functions increases, the cost and duration for optimal calibration become prohibitively large.  In order to reduce the number of dynamometer tests required for calibrating  modern engines, a large-scale simulation-driven machine learning approach is presented in this work.  A parallel, fast, robust, physics-based reduced-order engine simulator is used to 
obtain performance and emission characteristics of engines over a wide range of control parameters under various transient driving conditions (drive cycles). We scale the simulation up to 3,906 nodes of the Theta supercomputer at the Argonne Leadership Computing Facility to generate data required to train a machine learning model. The trained model is then used to predict various engine parameters of interest, and the results are compared with those predicted by the engine simulator. Our results show that a deep-neural-network-based surrogate model achieves high accuracy: Pearson product-moment correlation values larger than 0.99 and mean absolute percentage error within 1.07\% for various engine parameters such as \ext, \expr,  nitric oxide, and engine torque.  Once trained, the deep-neural-network-based surrogate model is  fast for inference: it requires about 16 microseconds for predicting the engine performance and emissions for a single design configuration  compared with about 0.5 second per configuration with the engine simulator. Moreover, we demonstrate that transfer learning and retraining can be leveraged to incrementally retrain the surrogate model to cope with new configurations that fall outside the training data space.

\keywords{Transient driving cycle modeling \and Surrogate modeling \and Machine learning  \and  Deep learning \and Deep neural networks}

\end{abstract}

\section{Introduction}
\input{intro}

\section{Surrogate modeling for transient drive cycle simulation}
\input{proposed}

\section{Experimental results}
\input{results}

\section{Related work}
\input{related-work}

\section{Conclusion}
We developed MaLTESE, a simulation-driven machine learning modeling framework that couples massively parallel simulations of thousands of engine drive cycles at real-time speeds and a machine-learning-based surrogate modeling approach. We scaled the simulation up to 3,906 nodes on the Theta supercomputer at the Argonne Leadership Computing Facility to generate data for training the surrogate model. We developed a deep-neural-network-based surrogate model and compared it with several classical machine learning methods.  
From our  numerical experiments we observed that all learning methods yielded reasonably good prediction accuracy. We demonstrated that the deep neural network is a promising method: it outperforms other classical machine learning techniques and achieves correlation coefficient values larger than 0.99 and a mean absolute percentage error within 1.07\% for \ext, \expr, nitric oxide, carbon monoxide, and \et.  Our comparative study of machine learning methods provides  valuable input to design engineers who can make an informed decision about the use of machine learning methods for their design and development assessments. 

In addition to the prediction accuracy of various machine learning methods, we studied the training and inference times for the various learning methods. We observed that the training time for deep neural networks was about two to four orders of magnitude higher than that for classical machine learning methods: 0.1--10 s for classical methods vs 1000 s for deep neural networks. Once the model is trained, however, the interference time required by the deep neural network to predict the output characteristics 6,000 data points (4 different drive cycles) is about 0.1 seconds (16 microseconds/configuration).  As shown earlier, concurrent simulation of four different drive cycles on four KNL processors would take over 700 seconds. These inference timing studies show that the deep-neural-network-based surrogate-model can be used for real-time control using the emerging low-cost and relatively low-powered on-board deep learning chips. 

The parametric study of the size of the training set showed that for predicting all output variables within 1\% accuracy, 48,000 data points (corresponding to 32 different representative drive cycles) were required.  This study shows that a small subset of well-chosen representative drive-cycles (64 drive cycles in this case) can be used to predict the output of other drive cycles without having to simulate the entire parametric range (250,000 drive cycles).  Based on the transfer learning studies, we have demonstrated the possibility of using machine learning methods to yield high-accuracy prediction even when the input parameter space is considerably different from the parameter range used for training.
\section*{Acknowledgment}
This research used resources of the Argonne Leadership Computing Facility, which is a DOE Office of Science User Facility supported under Contract DE-AC02-06CH11357. This material was based upon work supported by the U.S. Department of Energy, Office of Science, under Contract DE-AC02-06CH11357.

\bibliographystyle{splncs04}
\bibliography{bibs/references,bibs/pcmod,bibs/perfpred}

\begin{subappendices}
\renewcommand{\thesection}{\Alph{section}}%
\section{}
\input{appendix}
\end{subappendices}

\end{document}

%% file: intro.tex
In order to achieve the goals of increased fuel economy and performance while reducing emission, automotive manufacturers have implemented various strategies and parameter variables to control and optimize automotive engines. Engine calibration---the process of determining the optimal values of control variables such as spark/fuel injection timing, valve timing, exhaust gas recirculation (EGR) fraction) is of paramount importance in achieving high engine performance and fuel economy while meeting emission standards.
Currently, to make the problem tractable, automotive manufacturers optimize one or more engine performance indices (e.g., fuel economy, emissions, or engine torque) with respect to a given set of engine-controllable variables such as valve timing, EGR fraction, or ignition/injection timing, with all other conditions such as engine speed and load remaining the same. Optimal values of various engine operating points (speed and load) are obtained via dynamometer tests that are then used to generate engine maps.  This procedure is called static calibration for steady-state conditions. These static calibration values are then interpolated to obtain optimal operating conditions for other operating points.       
The static calibration process, however, presents significant and unique challenges on account of the large design space and conflicting constraints.  Over thirty independent design variables, including engine speed (i.e., RPM), torque, air-to-fuel ratio (AFR), and driving conditions (e.g., city or highway) influence the fuel economy, engine performance, and emissions.  Moreover, most engines are operated in transient mode, especially during city drives.  During the transient mode of operation, the engine speed and load change continuously and frequently (as opposed to a highway drive), and hence optimal operating conditions derived from static calibrations are not accurate.  The lack of accuracy stems from the fact that there is a strong nonlinear correlation between various input parameters and outputs. For instance, a small change in the spark timing can increase the engine torque but also greatly increase the NO emission.  In order to increase the accuracy for transient engine operation, more calibration tests have to be conducted over a wider range of input/controllable parameters to span the entire feasible engine operating domain. Hence, the cost and duration of the calibration process grow exponentially with the number of input/controllable parameters, greatly increasing the product design cycle/time to market.  Even for engines with simple technologies, achievement of the optimal calibrations for the transient driving mode is impractical.

Harnessing the power of high-performance computing, one can perform optimal calibrations for the transient driving conditions using massively parallel computations.  Conducting design, analyses, and optimization studies over such a large parameter space presents serious computational challenges, however.  To span the entire engine operating range over the vast parameter space requires thousands of combinations of input conditions.  For instance, if one were to consider just six different input control parameters with five parametric values for each input variable, one would have 15,625 ($5^6$) different input combinations for a single transient drive cycle (or commute of a single driver).  Given the wide variability in the driving habits of individual drivers and different types of commutes, simulating the typical drives of a handful of drivers would yield over a 100,000 transient simulations.  Each such simulation would produce vast amounts of output data, such as peak, average, and cumulative values of emissions, power, engine temperature, and exhaust gas temperature and pressure.  Computational time for a typical city or highway drive is also a major barrier to the use of high-performance computing in large-scale transient drive cycle simulations.  For instance, the computational time for a single engine cycle (one compression stroke followed by one expansion stroke of the piston) can range from a few hours to days at the strong-scaling limit (50--100 cores) of modern multidimensional simulation codes.  A typical 25--30-minute drive involves  about 40--50,000 engine cycles. Thus, a single multidimensional drive cycle simulation would require well over a year, which precludes their use for such drive cycle simulations and optimization (calibration).

Given the need to simulate typical drive cycles of thousands of vehicles in real time (physical time taken to run engine dynamometer tests or dyno tests) while efficiently harvesting and learning useful design, development, and optimization data, we have developed a modeling framework called MaLTESE\footnote{aptly named after a small, intelligent dog that loves to learn new tricks} (Machine Learning Tool for Engine Simulations and Experiments). It is a scalable simulation-driven machine learning (ML) framework that enables automotive design engineers to exploit the task parallelism inherent in simulating thousands of transient drive cycles and learning at real-time speeds. The framework also allows the coupling of experimental engine data in order to tune simulation constants and/or train the neural network and hence closely couples large-scale simulations, available engine data, and ML. This paper describes the use of MaLTESE to conduct the largest transient driving cycle simulation conducted on the Theta supercomputer at the Argonne Leadership Computing Facility. We also present  an in-depth study of the use of ML methods to predict engine performance and emissions based on the training and test data obtained from the drive cycle simulations.

The MaLTESE framework consists of two main components: an engine simulator pMODES and a neural-network-based surrogate-modeling tool.  Engine simulations of thousands of different typical transient city driving commutes, each approximately 25--30 minutes, were accomplished by using pMODES (parallel Multi-fuel Otto Diesel Engine Simulator). This  is a parallel, robust, physics-based real-time engine simulator that can concurrently compute the performance and emissions for thousands of transient drive cycles.  The simulator can perform engine simulations for either gasoline (Otto) or diesel engines with any combination of over thirty user-defined input/control variables.  Given a set of driving conditions (wind speed, friction, gear-shift/transmission strategy, etc), one can obtain detailed information about over twenty engine outputs, such as fuel consumption, engine performance (power/torque), emissions (carbon dioxide, carbon monoxide, nitric oxide, soot), exhaust gas temperatures and pressures, and maximum engine temperature and pressure.  The engine simulator produces the same data as an engine being tested on a dynamometer.  Since thousands of driving commutes can be simulated simultaneously, accurate input/output correlations (transient calibration) over a wide range of input parameters can be accomplished without the prohibitive testing costs.  Furthermore, since the drive cycle simulations can be conducted at speeds faster than real time, a typical drive cycle simulation can be conducted in less than 30 minutes, hence making it practical for the design and development of fleets of cars. A subset of the large calibration data is then input to the neural-network-based surrogate modeling tool.  Based on the calibration data, a surrogate model is trained to capture the relationship between the multiple inputs and outputs.  The trained surrogate model can then be used to predict expected calibration values of other driving conditions and can be a part of the engine control unit. Large computing clusters with thousands of cores greatly reduce the wall time and effort by concurrently simulating thousands of driving cycles. 
A subset of the large data set was generated from over 300 million engine operating points in a typical commute of 250,000 different drivers.  Finding the optimum operating condition for a given engine operation (speed, load, driving condition) can improve engine efficiency, reduce emissions, reduce engine wear and tear, and improve fuel economy. Use of large-scale computing and data analytics for drive cycle analyses enables engine designers to reduce the cost and time required for engine dyno tests,  hence reducing the product design cycle and cost to consumers.

The main goal of this paper is to use MaLTESE to demonstrate the following:
\begin{enumerate}
    \item Concurrent simulation of thousands of driving cycles with the engine simulator (pMODES) for a typical 25-minute commute at faster-than-real-time speed
    \item Ability of deep neural networks to use a small subset of the parameter space to train a model and predict engine output characteristics of any arbitrary driving cycle in the parameter space
    \item Inference time of a deep-neural-network-based surrogate model being considerably lower than simulations with near 1\% error in prediction accuracy.
\end{enumerate}
This paper is organized as follows. Section 2 describes the method of solution for the engine simulation and the training and testing of the neural-network-based ML predictions.  Section 3 presents the numerical experiments using various ML methods. Section 4 discusses related work.  The main conclusions of the paper are presented in Section 5.

%% file: proposed.tex
In this section, we discuss the engine simulator and the ML approach for surrogate modeling. We also describe the parameters of the drive cycle simulations and the choice of the parameter subspace to train the neural network. 

\subsection{Engine simulator}
\input{pmodes}

\subsection{ML-based surrogate modeling}
\input{surrogate}

%% file: pmodes.tex
The engine simulator pMODES \cite{Aithal13cst,Aithal13hcci} is used to compute the temporal variation of various engine parameters
such as pressure, temperature, and mixture composition for each CAD over an
entire drive cycle. The energy equation shown
in Eq.~(\ref{eqn:eneq}) describes the relationship between the engine
crank angle $\theta$ and instantaneous pressure ($P(\theta$)). 
\begin{equation}
\frac{ dP\left(\theta\right)}{d\theta}=\frac{\gamma-1}{V\left(\theta\right)}
\left(Q_{in}-Q_{loss}\right)
-\gamma\frac{P\left(\theta\right)}{V\left(\theta\right)}\frac{dV}{d\theta}
\label{eqn:eneq}
\end{equation}
Here, $Q_{in}$ is the heat input due to fuel combustion, $Q_{loss}$ is the heat lost from the engine, $\gamma$ is the ratio of specific heats of the working fluid, and $V(\theta$) is the instantaneous volume of the cylinder.  Solution of this equation yields the temporal
variation of cylinder pressure for a given set of operating conditions (such as
load, combustion duration, fuel type, and engine RPM).  The instantaneous
values of temperature and composition of the burned and unburned gas zones can be
obtained from the instantaneous value of computed pressure.  
Knowing the
instantaneous temperature, pressure, and
composition of the burned zone, one can compute  emissions such as nitric oxide, carbon monoxide, soot, and unburned hydrocarbons using simplified reduced 
chemistry models.  Details of these models and the solution procedure are 
discussed in
Ref.~\cite{SA13SAE}. Instantaneous values of equilibrium concentrations of the
combustion products are needed in order to compute various emissions.  Computation of
these equilibrium concentrations poses serious numerical challenges because of
the stiffness of the system of nonlinear equations describing the formation of
combustion products.  References~\cite{Aithal13cst,Aithal13hcci} discuss the details
of the computation procedure and steps taken to ensure a fast, robust solution.
Following the solution procedure discussed above,  one can obtain a temporal variation of output quantities such as emissions (NO, CO), engine exhaust temperature and pressure, and torque as a function of time.  Figure 1 in Ref. \cite{aithal2015accolades} shows the temporal variation of NO and CO for a given fuel injection pattern.

In this work we considered sixteen driving cycles. Each transient cycle had 1,500 data points corresponding to a typical 25-minute commute, with data sampled every second (25*60). For each drive cycle, we considered five values for six independent engine parameters---spark timing, engine rpm (depends on gear ratio), ambient air temperature, air humidity, internal EGR fraction (proportional to valve timing), and compression ratio (engine size)---thus yielding 15,625 cases ($5^6$) with different input conditions for each drive cycle and 250,000 for all  sixteen drive cycles considered. This number of 250,000 drive cycles is representative of the rush-hour traffic on four major freeways in a typical large city.

%% file: surrogate.tex
A class of ML approaches  used for surrogate modeling is supervised learning \cite{bishop2006pattern}. Typically, it is used to model the relationship between the output variables and several independent input variables. In this work, we seek to find a surrogate model that captures the relationship between the five output variables (\ext, \expr, \no, \co, and \et) and the ten input variables (ambient air temperature, air humidity, valve timing, engine size, spark timing,  gear ratio, fuel injection rate, air-fuel ratio, engine inlet pressure, and intake air mass). A supervised learning method takes as input a set ${\cal T}$ of $N$ training points of the form $\{(x_1, y_1), \ldots, (x_N, y_N)\}$, where $x_{i}$ and $y_{i}$ are the input and output vectors of the $i$th training point, respectively. The training procedure of the supervised learning method seeks to find a surrogate function $h$ for $f: X \to Y$, where $f$ is an unknown function that maps the multidimensional input space $X$ to the multidimensional output space $Y$, respectively, such that the difference between $f(x_i)$ and $h(x_i)$ is minimal for all $x_i \in {\cal T} \subset \cD$, where $\cD$ is the full data set.

Arguably, classical ML methods are limited in their ability to learn directly from raw data. 
For decades, the development of ML surrogate models required considerable domain expertise to 
transform raw input data into a suitable internal representation from which the system could try to learn the relationship between inputs and outputs. Recently, representation learning methods have been developed to automatically discover representations that are best for learning the relationship between inputs and outputs\cite{goodfellow2016deep}. Deep learning approaches \cite{lecun2015deep} are representation learning methods with multiple levels of representation. They are obtained by composing simple nonlinear computational units that transform the representation at one level into a representation at a higher, slightly more abstract level. These approaches have dramatically improved the state of the art in many ML tasks, such as speech recognition, visual object recognition, drug discovery, and genomics \cite{lecun2015deep,goodfellow2016deep}.

Deep neural network (\dnn) \cite{lecun2015deep} systems are a prominent class of deep learning approaches. A \dnn comprises a stack of computational layers organized in a hierarchical way, with the layers  connected through a system of weighted connections. Each layer has a number of simple computational units, each with a nonlinear transformation operation called an activation 
function. The input layer of the \dnn receives a batch of input data, which is transformed into higher-level representations through the stack of computational layers and weighted connections. The output layer of the \dnn gives the predicted values of the outputs. During the training phase, the weights of the connections in the network are adjusted to minimize prediction errors. This adjustment is achieved efficiently by using a back propagation method that calculates the gradient of the error with respect to all the weights in the network and uses it in a stochastic gradient-based optimization to adjust the connection weights.

While there exists a standard \dnn configuration for traditional ML tasks such as image and text classification, there is no default or general-purpose \dnn configuration for surrogate modeling of engineering applications and in particular transient drive cycle modeling. Designing a suitable \dnn for a given modeling task is a key research challenge for many nontraditional ML tasks. 

\begin{figure}[t]
        \includegraphics[width=\textwidth]{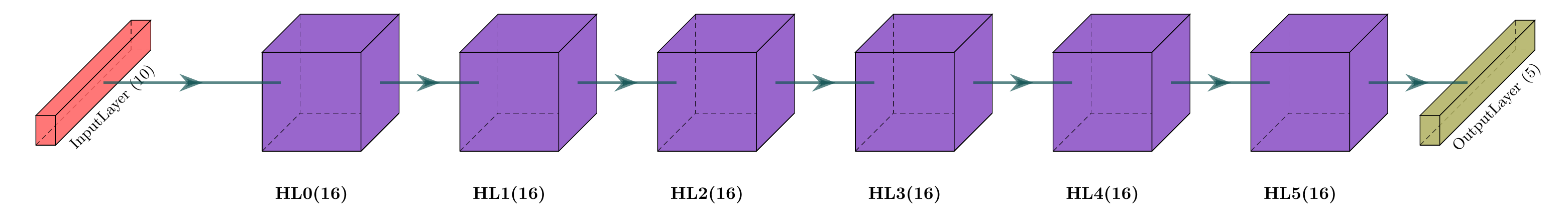}
        \vspace{-5mm}
\caption{The \dnn configuration obtained for transient drive cycle 
surrogate modeling}\label{fig:dnn}
\end{figure}

We carried out an exploratory study and developed a relatively simple multilayered feed-forward neural network. Figure \ref{fig:dnn} shows the obtained network used in this work: the input layer of size $|X|=10$ is connected to a dense hidden layer with $16$ units (HL0) and a rectified linear activation function (not shown in the figure). This configuration is repeated six times (HL1,$\ldots$, HL5), where the output of the previous layer is given as the input for the next layer. Consequently, the 16 units of the $j$th layer are connected to each of the 16 units in the $(j+1)$th layer. The last layer is the output layer of size $|Y|=5$ that gives predicted values.

%% file: results.tex
In  this  section, we first describe the  setup that we used to assess the efficacy of the proposed \dnn method. We then describe the training data generation and  prediction accuracy results.

\subsection{Setup}

In addition to \dnn, many classical ML methods (sometimes referred to as shallow learning methods) for surrogate modeling exist in the literature. 
Based on the algorithmic similarity and functionality, they can be grouped as regularization,
instance-based, recursive partitioning, kernel-based, bagging, and boosting methods. 
For comparison with \dnn, we selected several classical ML methods to cover different groups:
ridge regression (\rg) \cite{hoerl1970ridge},
$k$-nearest-neighbor regression (\knn)  \cite{bishop2006pattern},
support vector machine (\svm) \cite{smola2004tutorial},
decision tree (\dt) \cite{loh2011classification},
random forest (\rfr) \cite{randomForest},
extremely randomized trees (\etr) \cite{geurts2006extremely},
ADA-boosting regression (\abr) \cite{drucker1997improving}, 
bagging regression (\br) \cite{breiman1996bagging,louppe2012ensembles},
gradient boosting regression (\gbr) \cite{friedman2002stochastic}, and 
eXtreme gradient boosting (\xgb) \cite{chen2016xgboost}.
As a baseline, we also included the simplest regression method, 
multivariate linear regression (\lm).

The ML training and inference experiments were run on a single-node hardware platform with a 3.4 GHz Intel Xeon E5-2687W processor (8 cores per CPU), 64 GB RAM, with an NVIDIA Tesla P100, 16 GB GPU RAM. The \dnn training and inference leveraged GPUs, whereas the classical ML methods used only the host CPU processors.

We used Python (Intel distribution, version 3.6.3) and the scikit-learn library \cite{scikit-learn} (version 0.19.0) to implement all the classical ML methods. We used the default hyperparameters  provided by the scikit-learn library for the ML methods. For \dnn, we used Keras \cite{chollet2015keras} (version 2.0.8), a high-level neural network Python library that runs on the top of the TensorFlow library \cite{abadi2016tensorflow}  (version 1.3.0). We used the following hyperparameter settings for \dnn training: epochs=50, batch size=16, loss=mean squared error, and optimizer=adam. While \dnn natively supports multioutput regression, where we can build a single model with multiple outputs, the classical ML methods considered in our study do not support multioutput regression. Therefore, we built one model for each output. We leveraged the  \texttt{MultiOutputRegressor} interface in the scikit-learn library to build the multioutput regression models.

Given the different ranges for inputs and outputs, ML methods benefit from preprocessing the  training and the testing data set. For each input and output, we applied \texttt{MinMaxScaler} and \texttt{StandardScaler} transformations in the scikit-learn library. The former scales the values between 0 and 1, and the latter removes the mean and scales the values to unit variance. We applied the two transformations before training and applied the inverse of \texttt{StandardScaler} and \texttt{MinMaxScaler} transformations after inference so that evaluation metrics were computed on the original scale. Note that the inverse transformations are required only for the predicted output values.

We adopted two evaluation metrics to assess the accuracy of the ML models on the test data and to compare them. The first metric is the Pearson product-moment correlation coefficient ($r$), which we use to measure the strength of a linear association between observed and predicted values on the test data. This metric ranges from -1 to +1. A value of 1 indicates a perfect linear relationship between observed and predicted values. A value of 0 indicates that no linear correlation exists between observed and predicted values and thus the prediction accuracy of the model is poor. A value of less than 0 means that as the value of observed (predicted) values increases, the value of the predicted (observed) values decreases. While this metric does not capture the absolute error, it is particularly useful when engineers build ML models for optimization as an end goal, where the relative ordering of the predicted values is sufficient to choose the best configurations. The second metric is the mean absolute percentage error (MAPE) given by the mean of $100 \times \frac{|y^i - \hat{y}^i| }{y_i}$\% for $i \in {1,\ldots,n}$, where $y_i$ and $\hat{y}_i$ are observed and predicted values of the test data point $i$, respectively. We used this metric to assess the prediction error for each output.

\subsection{Training data generation at scale}

\input{data-generation}

\subsection{Comparison of ML methods}
\input{method-comparison}

\subsection{Impact of training set size}
\input{trainingsize}

\subsection{Model adaptation using transfer learning and retraining}
\input{transferlearning}

%% file: data-generation.tex
As explained earlier, 250,000 different transient drive cycles were simulated concurrently by using the engine simulator pMODES to generate the training and test data for the ML algorithms.   
The simulations were conducted on Theta---a 4,392-node, 11.69-petaflop Cray XC40--based leadership-class supercomputer at the Argonne Leadership Computing Facility (ALCF). Each node of Theta is a 64-core Intel Xeon Phi processor with 16 gigabytes of high-bandwidth in-package memory, 192 GB of DDR4 memory, and a 128 GB SSD. The nodes of Theta are interconnected by an Aries fabric. Theta has a total file system capacity of 10 petabytes.

In this work, large-scale computing was used to exploit the inherent task parallelism in the simulation of a large number of drive cycles. In such applications, it is important to demonstrate that the overall size of the problem (number of drive cycles considered) does not adversely affect the total wall time for simulation.  In order to test the weak-scaling characteristics of the simulation, three different tests were run, with 62,500, 12,5000, and 250,000 cases run concurrently, corresponding to $1/4$, $1/2$, and near-full-machine simulation (3,906 nodes out of 4,392 nodes). Since each drive cycle was run concurrently on a single processor, the total wall time for each of these cases should be nearly constant. Within each set of runs, the simulation time for an individual drive cycles depends on the computations required for the emissions, which can vary depending on the case being considered.  Table \ref{scaling} shows the minimum and maximum time required for computing an individual drive cycle.

\begin{table}[!hbtp]
\caption{Weak Scaling on Theta \label{scaling}}
\centering
\begin{tabular}{|c|c|c|}
\hline\hline
Nodes (cases)  & Min Time (sec) & Max Time (sec)\\
\hline
1024 (62500) & 728 & 1157  \\ 
\hline 
2048 (125000) &740 & 1252  \\
\hline
3906 (250000) &720 & 947 \\
\hline 
\hline 
\end{tabular}
\end{table}
The runtime for a typical 25-minute drive cycle was about 12--15 minutes (faster than real time) on the Intel Knights Landing (KNL)  cores on Theta. 
From Table \ref{scaling}, we can see that the minimum simulation time  is nearly constant for all the cases considered. We also can see that the maximum simulation time is nearly constant for the $1/4$ and $1/2$ machine size cases, whereas the near-full-machine simulation is about 20\% lower. The system load from other jobs on the machine seems to have a greater impact on the simulation time for cases 1 \& 2  compared with case 3, where there is less interference from other jobs on the system.  These results demonstrate that one can simulate drive cycles of various sizes---even thousands---on a large-scale production cluster such as Theta without a serious penalty on overall wall time for computation as the size of the drive cycle simulations increases.
Such a capability might be required if one were to use physics-based models to develop the acceleration and braking strategies for connected vehicles in order to optimize fuel efficiency, reduce emissions, and reduce engine wear and tear.  

The  discussion also shows that conducting a drive cycle simulation of a large parameter space requires considerable computational resources.  To minimize the use of large-scale computing for drive cycle analyses, we  investigated the possibility of using machine learning techniques wherein a small subset of the large parameter space is used as training data. We explored the possibility of using a trained model in predicting the characteristics of other drive cycles without the need for conducting simulations or gathering engine data with acceptable accuracy.  In this work, we generated the following data sets for training and testing.
\begin{enumerate}
    \item \trainone: From the complete parametric study of 250,000 different drive cycles, a representative set of 64 different drive cycles, spanning the input parameter range, was chosen for training data. We used Latin hypercube sampling \cite{osti_5236110}, a statistical method on the input space to select the 64 representative sets. Since each drive cycle had 1,500 data points for fuel flow rate, a total of 96,000 data points were used for training.
    \item \testone: To test the accuracy of prediction from the training set,  we used two different test data sets  from the entire set of 250,000 (excluding those used in the training).  The first set comprised four different drive cycles, which  had the same fuel flow rate but for which the other input parameters were different, for a total of 6,000 data points. 
    \item \testtwo:  In addition to this test data set, a  random set of four drive cycles (for a total of 6,000 points) was chosen for the 250,000 cases (excluding those used to train the model). By random, unlike the \testone case, no parameter was intentionally kept constant.
    \item \testthree: Both the  test cases \testone and \testtwo were drive cycles wherein the range of input parameters of the test drive cycles was the same as that for the training model. In order to test the efficacy of the ML methods wherein the test data might have parameters beyond the bounds of the trained data, a third data set was generated. This data set had a fuel flow rate that was 20\% higher than the corresponding fuel flow rate used in \testone.  Furthermore, the engine RPM was lower than that used in \testone by 17\%. 
\end{enumerate}

%% file: method-comparison.tex
\begin{figure}
    \centering
    \begin{subfigure}[b]{1.0\textwidth}
        \includegraphics[width=\textwidth]{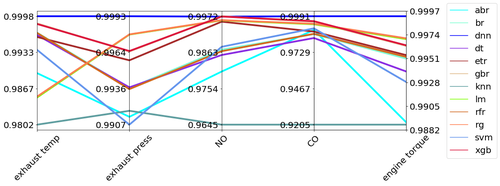}
        \vspace{-5mm}
        \caption{Pearson product-moment correlation coefficients on \testone}
        \label{fig:t1rho}
    \vspace{2mm}
    \end{subfigure}
    \begin{subfigure}[b]{1.0\textwidth}
        \includegraphics[width=\textwidth]{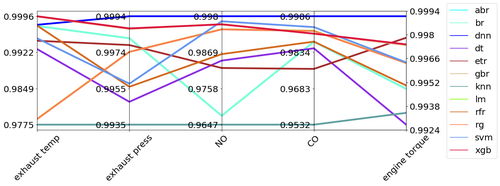}
        \vspace{-5mm}
        \caption{Pearson product-moment correlation coefficients on \testtwo}
        \label{fig:t2rho}
        \vspace{2mm}
    \end{subfigure}
    \begin{subfigure}[b]{1.0\textwidth}
        \includegraphics[width=\textwidth]{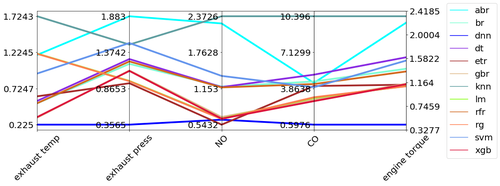}
        \vspace{-5mm}
        \caption{Mean absolute percentage errors (MAPE) on \testone}
        \label{fig:t1mape}
        \vspace{2mm}
    \end{subfigure}
    \begin{subfigure}[b]{1.0\textwidth}
        \includegraphics[width=\textwidth]{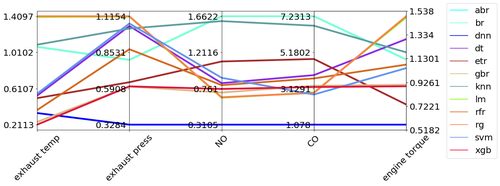}
        \vspace{-5mm}
        \caption{Mean absolute percentage errors (MAPE) on \testtwo}
        \label{fig:t2mape}
    \end{subfigure}
    \caption{Parallel coordinate plot showing the accuracy metrics obtained by different ML methods} \label{fig:ml-comparison}
\end{figure}

In this section, we compare the different ML methods that were trained on \trainone and tested on \testone and \testtwo. We use parallel coordinate plots to visualize the accuracy metrics obtained by the ML methods on the five outputs. In the plot, each output is given an axis; the five axes are placed parallel to each other. Each axis can have a different scale because each output can have a different range of values. Given an ML method, its accuracy value on each axis is connected and visualized through a line. 

The parallel coordinate plots for Pearson product-moment correlation coefficients are shown  in Figures \ref{fig:t1rho} and \ref{fig:t2rho}. On both testing data sets, all ML methods obtain correlation coefficients larger than 0.92. On \testone, \dnn outperforms other ML methods, obtaining correlation coefficient values larger than 0.99 for \ext, \expr, \no, \co, and \et, respectively. The trend is similar on \testtwo, where \dnn achieves larger correlation coefficient values than those of the classical ML methods. An exception is for \ext, where the correlation coefficient of \xgb is slightly larger than \dnn.

Figures \ref{fig:t1mape} and \ref{fig:t2mape} show the MAPE values on \testone and \testtwo, respectively. The range of error percentages for \ext, \expr, \no, and \et is between 0.2\% and 2.5\%; but for \co the error goes up to 10.39\% and 7.23\% on \testone and \testtwo, respectively. This indicates that prediction of \co is more difficult than prediction of \ext, \expr, \no, and \et. The MAPE  values obtained by \dnn are smaller than those of other ML methods. In particular, \dnn achieves significantly smaller MAPE values for the outputs; \expr, \co, and \et. Overall, MAPE values of \dnn are not more than 0.59\% and 1.07\% on \testone and \testtwo, respectively. The scatter plots of  observed  and  predicted  values from \dnn for the five outputs on \testone and \testtwo are given in Appendix A.

\begin{figure}
    \begin{subfigure}[b]{0.48\textwidth}
        \includegraphics[width=1.0\textwidth]{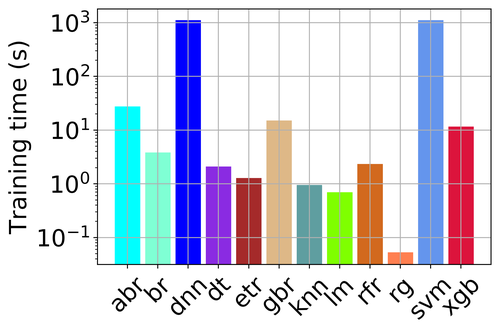}
        \vspace{-4mm}
        \caption{Time required for training different ML methods with \trainone (96000 training points)}
        \label{fig:training-time}
    \end{subfigure}
    \begin{subfigure}[b]{0.48\textwidth}
        \includegraphics[width=1.0\textwidth]{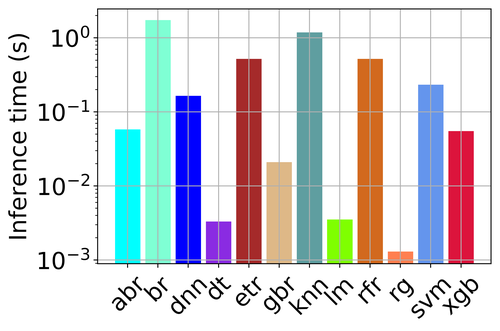}
        \vspace{-4mm}
        \caption{Time required for inference using different ML methods on \testone (6000 testing points)}
        \label{fig:inference-time}
    \end{subfigure}
    \caption{Bar plots showing the training and inference times of the different ML methods} \label{fig:ml-time}
\end{figure}

Figure \ref{fig:ml-time} shows the training and inference times of the different ML methods on \trainone and \testone, respectively. From \ref{fig:training-time}, we observe that \dnn requires approximately 1,000 seconds for training. On the other hand, the training times of classical ML methods range between 0.1 and 10 seconds. An exception is \svm, which requires a training time similar to that of \dnn. Even though \dnn leverages P100 GPUs, it is more computationally expensive than other ML methods. This difference can be attributed to the cubic algorithmic time complexity. The bagging and boosting methods typically have a time complexity of $O(N\ log\ N)$ in the training set size $N$. Figure \ref{fig:inference-time} shows the time required for inference on \testone. To predict 6,000 points, \dnn requires approximately 0.1 seconds (16 microseconds/configuration), which is lower than that of several sophisticated classical ML methods such as \br, \etr, \rfr, and \svm. Simple ML methods such as \lm, \rg, and \dt require less than 0.01 seconds, but their accuracy values are not as high. We observed a similar trend on \testtwo.

%% file: trainingsize.tex
\begin{figure}[t]
        \includegraphics[width=\textwidth]{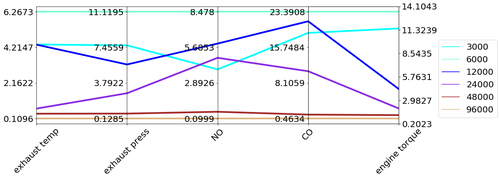}
        \vspace{-5mm}
    \caption{Parallel coordinate plot showing the impact of training data set size on the MAPE values obtained by \dnn on \testone}\label{fig:size}
\end{figure}

We studied the impact of the training data size on the accuracy of the \dnn method by varying the number of training points. In addition to the default training data size of 96,000, we considered training set sizes of 1,500, 3,000, 6,000, 12,000, 24,000 and 48,000 data points (1,500 data points represent one complete transient drive cycle; hence, the training set sizes represent 1, 2, 4, 8, 16, and 32 different drive cycles). For each training set size, we trained the \dnn method and evaluated the model on \testone. 

The results are shown in Figure \ref{fig:size}, where the lines in the parallel coordinate plot correspond to the training data set sizes. We observe that an increase in the training set size decreases the MAPE values. We did not include the MAPE values for 1,500 because the error values are too high (greater than 100\%), which results in skewed axes ranges. The \dnn model trained with 3,000 points yields MAPE values between 6.26\% and 23.39\%. While the MAPE values for training set size to 6,000 and 12,000 are lower than that of 3,000, to achieve MAPE values within 1\% for all the outputs, the \dnn model requires at least 48,000 training points.

%% file: transferlearning.tex
\begin{figure}[t]
    \centering
    \begin{subfigure}[b]{1.0\textwidth}
        \includegraphics[width=\textwidth]{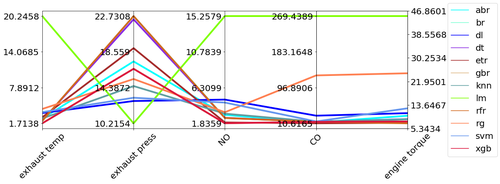}
        \vspace{-5mm}
        \caption{MAPE values obtained by the ML methods on \testthree}
        \label{fig:mag}
    \vspace{5mm}
    \end{subfigure}
    \begin{subfigure}[b]{1.0\textwidth}
        \includegraphics[width=\textwidth]{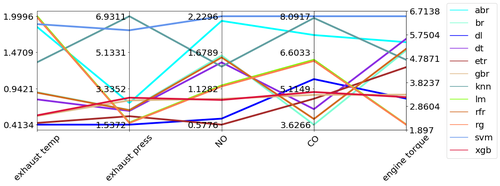}
        \vspace{-5mm}
        \caption{MAPE values obtained by the ML methods on \testthree with retraining and transfer learning}
        \label{fig:mag-transfer}
    \end{subfigure}
    \caption{Parallel coordinate plots showing the impact of retraining and transfer learning on the ML methods}\label{fig:retrain}
\end{figure}

We evaluated the efficacy of the ML models when the test data falls outside the training data regime. We took the ML models trained on \trainone and tested them on \testthree. The results are shown in Figure \ref{fig:mag}. The ranges of MAPE values for all ML models are  large: [1.71\%, 20.24\%] for \ext, [10.21\%, 22.73\%] for \expr, [1.83\%, 15.25\%] for \no, [10.61\%, 269.43\%] for \co, [5.34\%, 46.86\%] for \et. This range can be attributed to the fact that while ML methods can generalize the learned functional relationship inside the input space spanned by the training points, outside that space their prediction power decreases  significantly.

A promising approach to adapt ML models for new test cases such as \testthree involves calibrating the trained model by using transfer learning, where a model trained  on  one  task  can  be adapted  to  a  similar task  with  limited  training  data. In our case, a small subset of data from \testthree can be used to retrain the model. Nevertheless, not all ML methods offer that transfer learning capability. Among the ML methods considered in our study only \dnn can be used for transfer learning. All other methods require complete retraining, where one needs to  add the new data to the training data and train from scratch. 

We used 1,500 points from \testthree for transfer learning and retraining from scratch. We note that that the training data set size of 1,500 points alone resulted in poor prediction accuracy on \trainone. Our hypothesis is that using the ideas of transfer learning and retraining but with the same limited data setting, we can significantly improve the prediction accuracy of the ML models.

For \dnn, we took the model trained on \trainone. To enable transfer learning, we froze the  weights of  HiddenLayer0, HiddenLayer1, and HiddenLayer2 layers (see Figure \ref{fig:dnn});  used 1,500 points from \testthree; and  retrained  the \dnn model, where  the  weights of HiddenLayer3, HiddenLayer4, and HiddenLayer5 layers were adjusted. The retrained \dnn model was then used to predict the outputs in \testthree. For other ML models, we used the retraining-from-scratch approach, where we added 1,500 points from \testthree to 96,000 points of \trainone and trained the ML models.

The results are shown in Figure \ref{fig:mag-transfer}. We observe that both the transfer learning method and training from scratch for other ML methods significantly reduce error values for all the outputs. The ranges of MAPE values are [0.41\%, 1.99\%] for \ext, [1.53\%, 6.93\%] for \expr, [0.57\%, 2.22\%] for \no, [3.62\%, 8.09\%] for \co, and [1.89\%, 6.71\%] for \et. The \dnn method  obtains smaller MAPE values for three outputs: 0.41\% for \ext, 1.53\% for \expr, and 0.58\% for \no. Only for \co and \et are the MAPE values  larger. Overall, however,  MAPE is within 5.5\%.

%% file: related-work.tex
Given the importance of internal combustion (IC) engines in transportation and power generation,  considerable work has been conducted in the area of predicting their performance and emissions, including the use of soft computing techniques such as artificial neutral networks (ANNs).  Some of the earliest attempts to use ANNs to predict the \no emissions during the transient operation of a diesel engine were reported by Ref. \cite{krijnsen1999prediction}.  Similar attempts were reported in Refs. \cite{hashemi2007artificial} and \cite{parlak2006application}.  Reference \cite{shrivastava2018application} provides an exhaustive review of the use of soft computing techniques in automotive engines.  All the reported results with these techniques for IC engines are for a single engine/transient operation.  Most of these reported results use experimental data from an engine operation with a small set of input parameters (two to three input variables) to predict a specific output quantity such as NO or exhaust temperature.  None of the  papers have attempted to use machine learning techniques to predict the performance and emissions of a fleet of cars with a large set of input parameters, each of which was varied over a large operating range, as reported in this work.  

To the best of our knowledge, this is the single largest drive cycle simulation (250,000 different cases) conducted by using a well-validated, physics-based reduced-order model at faster-than-real-time computing speeds. This is also the first demonstration of the  ability to apply ML methods to such large-scale engine data to predict performance and emissions.

%% file: appendix.tex
\begin{figure}
    \begin{subfigure}[b]{0.48\textwidth}
        \includegraphics[width=1.0\textwidth]{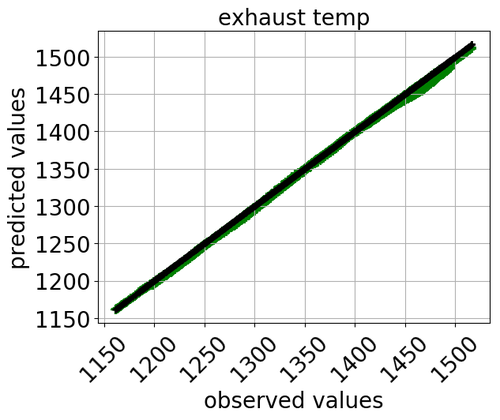}
        \vspace{-4mm}
        \caption{\testone}
    \end{subfigure}
    \begin{subfigure}[b]{0.48\textwidth}
        \includegraphics[width=1.0\textwidth]{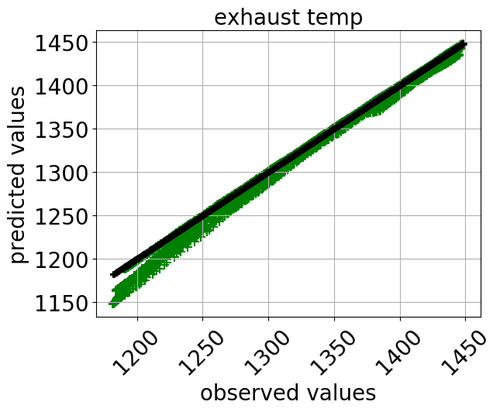}
        \vspace{-4mm}
        \caption{\testtwo}
    \end{subfigure}
    \caption{Scatter plot of observed and predicted values from \dnn for \ext on \testone and \testtwo} \label{fig:ext}
    \vspace{-15mm}
\end{figure}
\begin{figure}
    \begin{subfigure}[b]{0.48\textwidth}
        \includegraphics[width=1.0\textwidth]{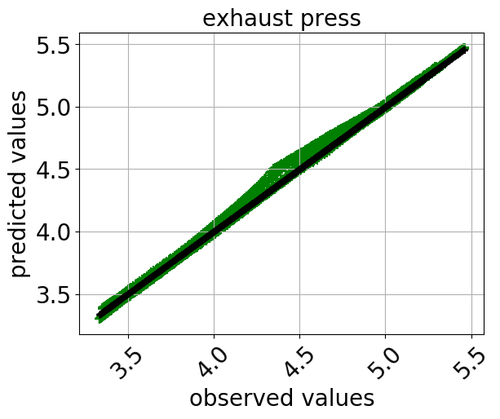}
        \vspace{-4mm}
        \caption{\testone}
    \end{subfigure}
    \begin{subfigure}[b]{0.48\textwidth}
        \includegraphics[width=1.0\textwidth]{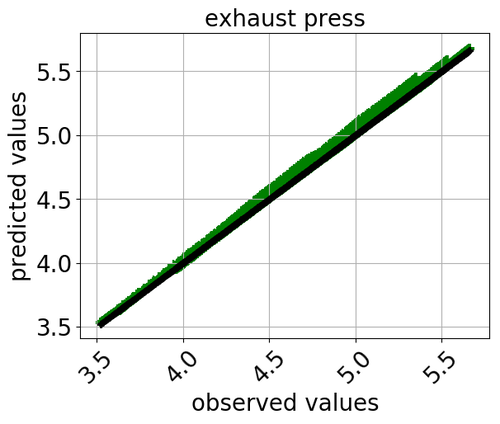}
        \vspace{-4mm}
        \caption{\testtwo}
    \end{subfigure}
    \caption{Scatter plot of observed and predicted values from \dnn for \expr on \testone and \testtwo} \label{fig:expr}
    \vspace{-20mm}
\end{figure}
\begin{figure}
    \begin{subfigure}[b]{0.48\textwidth}
        \includegraphics[width=1.0\textwidth]{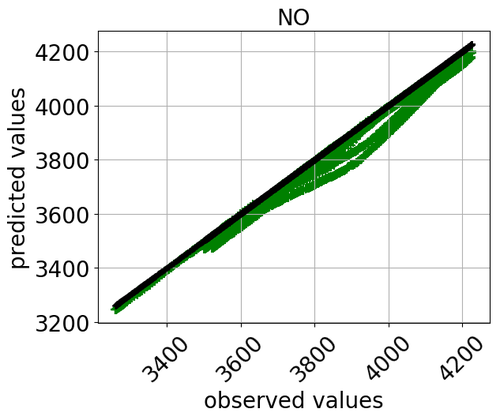}
        \vspace{-4mm}
        \caption{\testone}
    \end{subfigure}
    \begin{subfigure}[b]{0.48\textwidth}
        \includegraphics[width=1.0\textwidth]{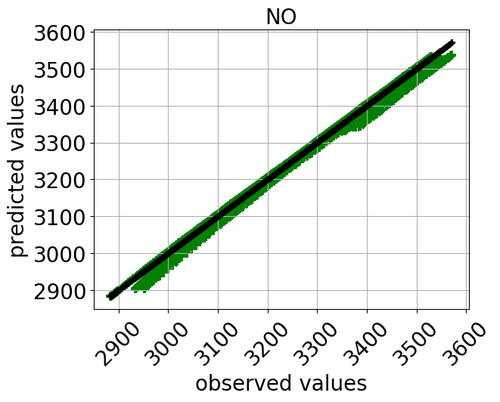}
        \vspace{-4mm}
        \caption{\testtwo}
    \end{subfigure}
    \caption{Scatter plot of observed and predicted values from \dnn for \no on \testone and \testtwo} \label{fig:no}
    \vspace{-15mm}
\end{figure}
\begin{figure}
    \begin{subfigure}[b]{0.48\textwidth}
        \includegraphics[width=1.0\textwidth]{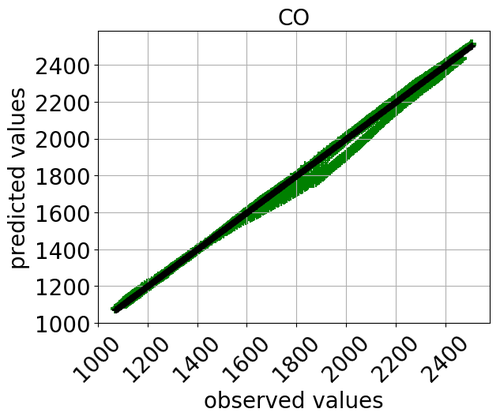}
        \vspace{-4mm}
        \caption{\testone}
    \end{subfigure}
    \begin{subfigure}[b]{0.48\textwidth}
        \includegraphics[width=1.0\textwidth]{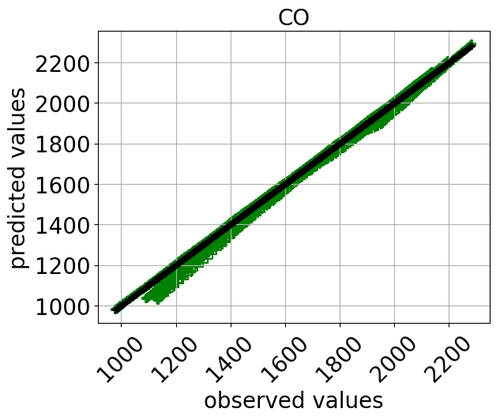}
        \vspace{-4mm}
        \caption{\testtwo}
    \end{subfigure}
    \caption{Scatter plot of observed and predicted values from \dnn for \co on \testone and \testtwo} \label{fig:co}
    \vspace{-15mm}
\end{figure}
\begin{figure}
    \begin{subfigure}[b]{0.48\textwidth}
        \includegraphics[width=1.0\textwidth]{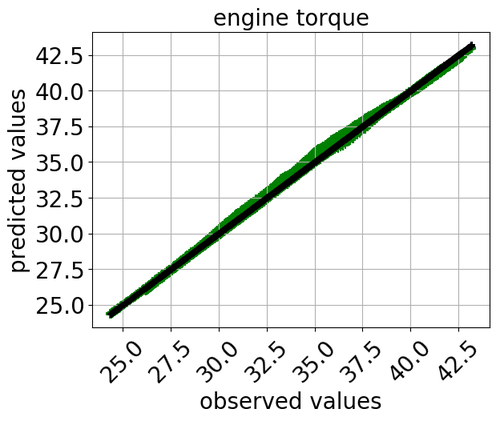}
        \vspace{-4mm}
        \caption{\testone}
    \end{subfigure}
    \begin{subfigure}[b]{0.48\textwidth}
        \includegraphics[width=1.0\textwidth]{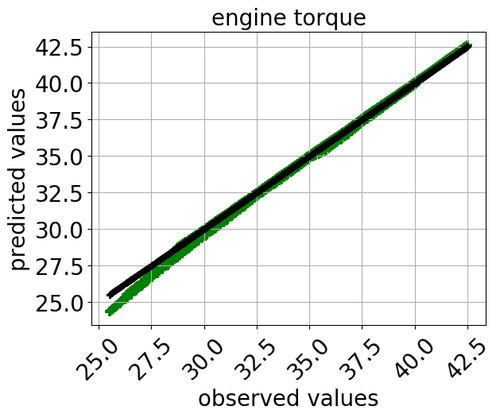}
        \vspace{-4mm}
        \caption{\testtwo}
    \end{subfigure}    
    \caption{Scatter plot of observed and predicted values from \dnn for \et on \testone and \testtwo} \label{fig:co}
    \vspace{-20mm}
\end{figure} 